\documentclass[10pt,twocolumn,letterpaper]{article}

\usepackage{iccv}              

\definecolor{iccvblue}{rgb}{0.21,0.49,0.74}
\usepackage[pagebackref,breaklinks,colorlinks,allcolors=iccvblue]{hyperref}
\usepackage{algorithm,algorithmic}
\usepackage{multirow} 
\usepackage{svg}
\usepackage{xcolor}
\usepackage{booktabs}
\usepackage{siunitx}
\usepackage{caption}
\usepackage{tabularx}
\usepackage{indentfirst}
\usepackage[accsupp]{axessibility}
\newcommand{\name}[0]{ViT-EnsembleAttack }

\def\paperID{4802} 
\def\confName{ICCV}
\def\confYear{2025}

\title{ViT-EnsembleAttack: Augmenting Ensemble Models for Stronger Adversarial Transferability in Vision Transformers}

\author{
Hanwen Cao\footnotemark[1] , Haobo Lu\footnotemark[1] ,
Xiaosen Wang, Kun He\footnotemark[2]
\\
School of Computer Science and Technology\\ Huazhong University of Science and Technology\\
{\tt\small \{hanwen,haobo,brooklet60\}@hust.edu.cn, xswanghuster@gmail.com}
}

\definecolor{darkyellow}{RGB}{204, 153, 0}
\begin{document}

\maketitle
\footnotetext[1]{The first two authors contributed equally.}
\footnotetext[2]{Corresponding author.}
\begin{abstract}
Ensemble-based attacks have been proven to be effective in enhancing adversarial transferability by aggregating the outputs of models with various architectures. 
However, existing research primarily focuses on refining ensemble weights or optimizing the ensemble path, overlooking the
exploration of ensemble models to enhance the transferability of adversarial attacks. 
To address this gap, we propose applying adversarial augmentation to the surrogate models,
aiming to boost overall generalization of ensemble models and reduce the risk of adversarial overfitting.
Meanwhile, observing that ensemble Vision Transformers (ViTs) gain less attention, we propose ViT-EnsembleAttack based on the idea of model adversarial augmentation, the first ensemble-based attack method tailored for ViTs to the best of our knowledge.
Our approach generates augmented models for each surrogate ViT using three strategies: Multi-head dropping, Attention score scaling, and MLP feature mixing, with the associated parameters optimized by Bayesian optimization.
These adversarially augmented models are ensembled to generate adversarial examples.
Furthermore, we introduce Automatic Reweighting and Step Size Enlargement modules to boost transferability.
Extensive experiments demonstrate that \name significantly enhances the adversarial transferability of ensemble-based attacks on ViTs, outperforming existing methods by a substantial margin. Code is available at \url{https://github.com/Trustworthy-AI-Group/TransferAttack}.

\end{abstract}


    
\section{Introduction}
Deep Neural Networks (DNNs), including Convolutional Neural Networks (CNNs)~\cite{he2016deep} and Vision Transformers (ViTs)~\cite{DBLP:journals/corr/abs-2010-11929}, are inherently vulnerable to adversarial attacks~\cite{goodfellow2014explaining,wei2022towards}, despite their impressive performance in solving various computer vision tasks.
Adversarial examples, carefully designed to deceive DNNs, can be transferred between different models~\cite{liu2016delving,wang2021boosting}, which means that a perturbation generated on a surrogate model can also mislead other models, even those with different architectures. 
This transferability enables a type of adversarial attack known as transfer-based attacks. 
Transfer-based adversarial examples are trained on surrogate models and can effectively attack unknown target models.
To mitigate the gap between surrogate models and target models, recent researches~\cite {wang2024boosting,lin2024boosting,li2024improving, wang2021enhancing,zhang2024bag} have introduced various techniques to improve transferability, such as input transformations~\cite{wang2023structure,lin2024boosting,ge2023improving} and advanced objective functions~\cite{li2024improving,zhang2022improving}.

\begin{figure*}[t]
\centering
\includegraphics[width=2.1\columnwidth]{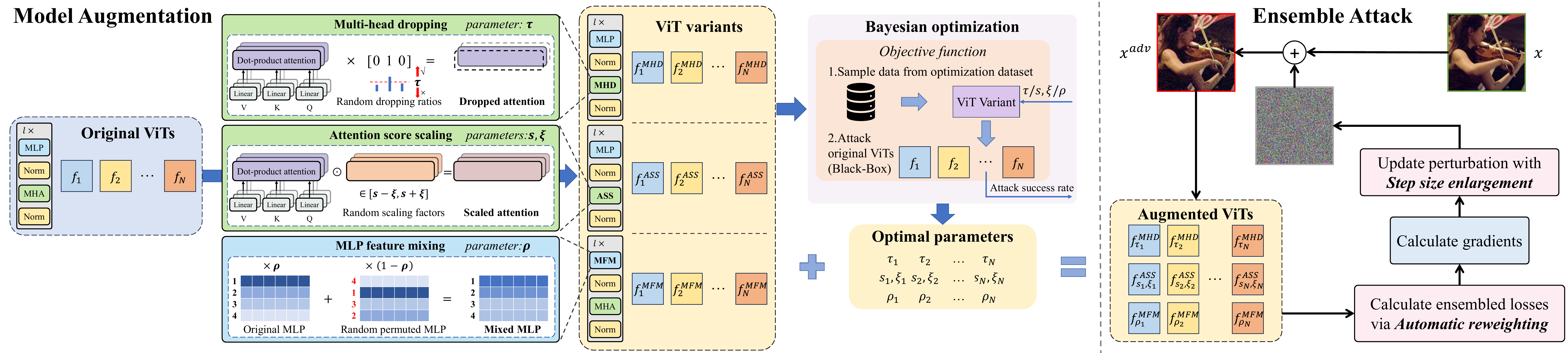}
\caption{Overview of the proposed ViT-EnsembleAttack framework.
The models $f_{1},...,f_{N}$ represent the $N$ original surrogate ViTs.
Unlike traditional ensemble-based attacks, ViT-EnsembleAttack generates a set of augmented models using three strategies with parameters optimized by Bayesian optimization, and ensembles these augmented models to produce adversarial examples. 
}
\label{baseline}
\end{figure*}
Ensemble-based attacks~\cite{liu2016delving} combine the outputs of multiple surrogate models to generate adversarial examples.
These attacks can be easily integrated with existing transfer-based methods, such as gradient-based MI-FGSM~\cite{dong2018boosting} or NI-FGSM~\cite{lin2019nesterov}, and input transformation methods like TI-FGSM~\cite{dong2019evading}, to further enhance attack performance.
Earlier approaches~\cite{liu2016delving} simply average the outputs of ensemble models, yielding modest transferability. 
Subsequent work has focused on reducing discrepancies among surrogate models and adjusting ensemble weights.
For instance, Stochastic Variance Reduced Ensemble adversarial attack (SVRE)~\cite{xiong2022stochastic} utilizes the idea of Stochastic Variance Reduced Gradient (SVRG)~\cite{johnson2013accelerating} to reduce the variances of gradient updates; 
Adaptive Model Ensemble Adversarial Attack (AdaEA)~\cite{chen2023adaptive} and Stochastic Mini-batch black-box attack with Ensemble Reweighting (SMER)~\cite{tang2024ensemble} dynamically adjust model weights based on adversarial contribution.\par
These methods have enhanced transferability by optimizing the combination of fixed surrogate models. 
However, we think it is not enough to merely focus on how to optimize the combination.
Prior works don't investigate the potential contributions of surrogate models themselves in enhancing attack transferability.
In other words, original surrogate models may not be the most effective surrogates for ensemble-based attacks. 
This gap motivates our approach of augmenting ensemble models adversarially. 
It is noteworthy that model augmentation can be achieved through various approaches.
Our approach focuses on increasing model diversity by introducing randomness into the model inference process.
This method requires designing randomization strategies tailored to the characteristics of the models and, more importantly, confirming the optimal degree of randomness.
In ensemble-based attacks, where multiple surrogate models are available, we can apply this augmentation to each individual surrogate. 
We then treat the others as black-box models to evaluate the transferability of the augmented model.
Higher transferability indicates a more suitable degree of randomness.
By doing this, all of the augmented surrogates can generate more diverse backpropagation paths for the same input than original surrogates, guiding the update of perturbations and thereby reducing the risk of adversarial overfitting.
\par
Given the superior performance of ViTs over CNNs in many tasks, we focus on designing an attack framework specifically for ViTs, which is less explored in existing works.  
We propose a novel ensemble-based attack, termed ViT-EnsembleAttack, against ViTs from the perspective of adversarially augmenting the ensemble models. 
Specifically, we draw inspiration from three data augmentation strategies—masking, scaling, and mixup—and propose three corresponding augmentation strategies for ViTs: Multi-head dropping (MHD), Attention score scaling (ASS), and MLP feature mixing (MFM).
Each original surrogate ViT will be modified through these strategies and generate three variants.
These variants are parameterized and will be optimized by Bayesian optimization to become augmented ViTs, which will be used as new surrogate models.
Additionally, we propose Automatic Reweighting to adjust the ensemble weights dynamically and Step Size Enlargement to accelerate convergence during the attack. 
The overview of \name is illustrated in Figure~\ref{baseline}.\par


The main contributions of this work are as follows:
\begin{itemize}
    \item We introduce a novel perspective to improve ensemble-based attack transferability by adversarially augmenting the surrogate models and propose, to the best of our knowledge, the first ensemble-based attack tailored for ViTs.
    \item We design three augmentation strategies tailored to the structure of ViTs and utilize Bayesian optimization to fine-tune the optimal parameters. 
    We further introduce Automatic Reweighting and Step Size Enlargement to improve the attack's efficiency.
    \item Comprehensive experiments validate the superior performance of \name in enhancing the adversarial transferability. Notably, our approach outperforms the state-of-the-art baseline by a clear margin of 15.3\% attack success rate on average when attacking CNNs.
\end{itemize}
\section{Related Work}
\subsection{Adversarial Attacks}
\textbf{Gradient-based attacks.}
Adversarial attacks differ from standard gradient descent, as they typically employ gradient ascent to reverse the optimization effect.
Goodfellow \etal~\cite{goodfellow2014explaining} introduced the Fast Gradient Sign Method (FGSM), which generates adversarial perturbation in a single step.
Based on this, Kurakin \etal~\cite{kurakin2018adversarial} and Dong \etal~\cite{dong2018boosting} proposed iterative versions of FGSM, the latter introducing momentum to stabilize the update direction. 
Although these methods achieve high performance in white-box settings, they struggle to maintain the same transferability in black-box settings, where information about the target model is typically unavailable. 

\textbf{Transfer-based attacks.}
Several approaches have been explored to improve adversarial transferability~\cite{liu2016delving,dong2018boosting,zhang2023improving,ge2023boosting}.
Xie \etal~\cite{xie2019improving} and Lin \etal~\cite{lin2019nesterov} combined the gradients of the augmented examples using resizing and scaling techniques to create diverse input patterns for higher transferability.
Ganeshan \etal~\cite{ganeshan2019fda} disrupted the deep features within DNNs, while Zhang \etal~\cite{zhang2022improving} extended this idea by calculating feature importance for each neuron.
Li \etal~\cite{li2020learning} targets ghost networks generated through aggressive dropout applied to intermediate features, and Wang \etal~\cite{xiaosen2023rethinking} mitigated gradient truncation by recovering gradients lost due to non-linear activation functions.
Although transfer-based attacks show promising performance in enhancing adversarial transferability between CNNs, their attack success rate diminishes when transferring to ViTs, which are known to exhibit greater robustness~\cite{wei2022towards}.

\textbf{Ensemble-based attacks.}
Ensemble-based methods fuse outputs of multiple models to enhance the effectiveness of transfer-based attacks.
Among the three common ensemble approaches, $i.e.$ ensemble on predictions, ensemble on losses, and ensemble on logits, Dong \etal~\cite{dong2019evading} showed that the latter is the most effective. 
Xiong \etal~\cite{xiong2022stochastic} proposed the SVRE method to reduce the variance among the ensemble models utilizing the idea of SVRG~\cite{johnson2013accelerating}  method. 
Chen \etal~\cite{chen2023adaptive} introduced AdaEA, which adaptively adjusts the contribution of each model in the ensemble and synchronizes update directions through a disparity-reduced filter, aiming to bridge the gap between CNNs and ViTs. 
Tang \etal~\cite{tang2024ensemble} proposed SMER, which generates stochastic mini-batch perturbations to enhance ensemble diversity and utilizes reinforcement learning to adjust ensemble weights.
In contrast, \name focuses on optimizing the surrogate models themselves rather than the ensemble path, by exploiting unique augmentations specific to ViTs.

\subsection{Adversarial Defenses}
Various approaches have been proposed to defend against adversarial attacks and improve the robustness of DNNs.
Adversarial training~\cite{tramer2017ensemble} is one of the most effective techniques, where clean images and their corresponding adversarial examples are incorporated into the training process.
Another category of adversarial defense focuses on input transformation techniques, which disrupt the adversarial pattern by preprocessing the input data. 
Popular methods in this category include reversing adversarial features~\cite{naseer2020self}, randomly resizing~\cite{xie2017mitigating}, utilizing compression techniques~\cite{guo2017countering}, and purifying inputs with  GANs~\cite{naseer2020self} or diffusion models~\cite{wang2023better}.
In this work, we select some defensive models as target models to assess the effectiveness of the proposed \name compared to existing SOTA baselines.
\section{Methodology}
\subsection{Preliminaries}
Given a clean image $x$ with the ground-truth label $y$, a surrogate ViT model $f$, the goal of the adversarial attack is to generate an adversarial image $x^{adv}=x+\delta$ to mislead the model $f$, i.e., $f(x^{adv})\neq f(x)=y$, where $\delta$ is the additive perturbation. 
A set of boundary conditions are imposed on the perturbation to make it imperceptible in relation to the clean example, $i.e.$ $\Vert \delta \Vert_{p}<\epsilon$, where $\Vert \cdot \Vert_{p}$ represents the $L_{p}$ norm.
To align with previous works, we employ $p=\infty$ for the following comparisons.
Therefore, the iterative attack process on a single surrogate model can be described as:
 \begin{equation}\label{I-FGSM} 
 x^{adv}_{t+1}=x^{adv}_t+\alpha \cdot \text{sign}(\nabla_{x^{adv}_t}J(f(x^{adv}_t),y)),
\end{equation}
where $\alpha$ is step size, $J$ is the loss function, sign(·) denotes the sign function, $x^{adv}_t$ denotes the adversarial example in $t^{th}$ iteration and $\nabla_{x^{adv}_t}J(f(x^{adv}_t),y)$ is the gradient of the loss function $w.r.t.$ $x^{adv}_t$.\par

Ensemble-based attacks utilize the output of multiple surrogate models and usually average them to obtain loss.
Assuming that there are $N$ surrogate models, the generation process of adversarial examples can be described as:
 \begin{equation}\label{Ensemble-I-FGSM} 
 x^{adv}_{t+1}=x^{adv}_t+\alpha \cdot \text{sign}(\sum_{i=1}^{N}w_i \cdot \nabla_{x^{adv}_t}J(f_{i}(x^{adv}_t),y)),
\end{equation}
where $w_i\geq0$ is the ensemble weight of each ensemble model $f_i$ and satisfies $\sum_{i=1}^{N}w_i=1$.\par
\subsection{Motivation}
Since the effectiveness of transferable adversarial attacks has been shown to be highly correlated with the diversity of the model~\cite{li2020learning,chen2023adaptive}, we argue that ensemble models can be adversarially augmented to be more diverse, thus further enhancing their adversarial transferability.
This inspires us to treat the ensemble models as tunable components, rather than fixed components as assumed in other studies.
Following this principle, we introduce ViT-EnsembleAttack, the first ensemble-based attack method tailored for ViTs to the best of our knowledge.


\subsection{The \name Method}
The \name method consists of three modules: Model Augmentation, Automatic Reweighting, and Step Size Enlargement. 
Detailed descriptions of these modules are provided below.
\par
\textbf{Model Augmentation.}
A typical ViT model consists of alternating layers of multi-head self-attention (MSA) and multi-layer perceptron (MLP) blocks.
To augment surrogate ViTs, we adapt three data-augmentation-inspired strategies on these special modules,
namely \textbf{\textit{Multi-head dropping}}, \textbf{\textit{Attention score scaling}}, and \textbf{\textit{MLP feature mixing}}.
We also design 
\textbf{\textit{Parameter optimization}} process to identify the optimal parameters.
Detailed descriptions are provided below.\par
\textit{Multi-head dropping (MHD)} means randomly abandoning some heads in each MSA.
In practice, we set a threshold $\tau \in [0,1]$ to determine whether to drop the head. 
Each head in each MSA of the surrogate ViTs will be independently assigned a random probability from 0 to 1 following a uniform distribution.
Heads with lower probabilities than $\tau$ will be dropped, \ie, the attention score matrix in this head becomes an all-zero matrix.
Here $\tau$ is the corresponding parameter to be optimized.
\par
\textit{Attention score scaling (ASS)} means that for each attention score matrix, we generate a matrix with random scaling factors $\in [s-\xi,s+\xi]$ following a uniform contribution. 
The scaling matrix has the same shape with the attention score matrix to make element-wise multiplication.
Here $s,\xi$ are the corresponding parameters to be optimized. 
\par 
\textit{MLP feature mixing (MFM)} randomly permutates the feature representations of MLP to form a new matrix.
Then mix the vanilla MLP matrix with $(1-\rho)$ and the new matrix with $\rho$ as the final output.
Here $\rho$ is the parameter to be optimized.
\par
\textit{Parameter optimization.}
Each surrogate model $f_i$ can generate three variants $f^c_{i,p_i}$ with the above strategies, where $c \in \{MHD,ASS,MFM\}$ means the augment strategy, $p_i \in \{\tau_i ,(s_i,\xi_i),\rho_i \}$ means the corresponding parameter(s).
For simplicity, we use $f^c_{p_i}$ in place of $f^c_{i,p_i}$.
We employ Bayesian optimization to optimize parameters for these variants.
The most important aspect of Bayesian optimization is a well-designed objective function that guides the search process.
In our method, we generate adversarial examples on $f^c_{p_i}$ and attack the other original surrogates $\{f_1,...,f_{i-1},f_{i+1},...,f_N\}$.
The average attack success rate on target models is set as the output of objective function, with the purpose of enhancing the transferability of the select model $f^c_{p_i}$.
Details of the objective function are listed in Algorithm \ref{OF}.
For convenience, we use $gp\_minimize$ function in Python library $skopt$ to build this Bayesian optimization process.
We denote the number of calls to the objective function as $n_{calls}$, the parameter selection space as $P$, and the remaining parameters of $gp\_minimize$ are set as default.

\begin{algorithm}[t]
\caption{Objective function for Bayesian optimization}
\label{OF}
\textbf{Input}: Parameter(s) $p$, augmentation strategy $c $, surrogate model $f$, test models set $F=\{f_1,...,f_{N-1}\}$, images for Bayesian optimization $X^B$ with corresponding ground-truth label $Y^B$, the number of randomly sampled images $M$.
\\
\textbf{Output}: Average attack success rate.

\begin{algorithmic}[1] 
\STATE Random choose $M$ images from $X^B$ and their corresponding labels to compose the attack datasets.\\
\STATE Modify $f$ to $f^c_p$ according to $c$ and $p$.\\
\STATE Using MI-FGSM algorithm generate adversarial examples $ \{x_1^{adv},...,x_M^{adv} \}$ on  $f^{c}_p$.\\
\STATE Calculate the average attack success rate of $ \{x_1^{adv},...,x_M^{adv} \} $ on test models $F$ .\\
\RETURN Average attack success rate.\\
\end{algorithmic}
\end{algorithm}

\begin{figure}[t]
\centering
\includegraphics[width=1\columnwidth]{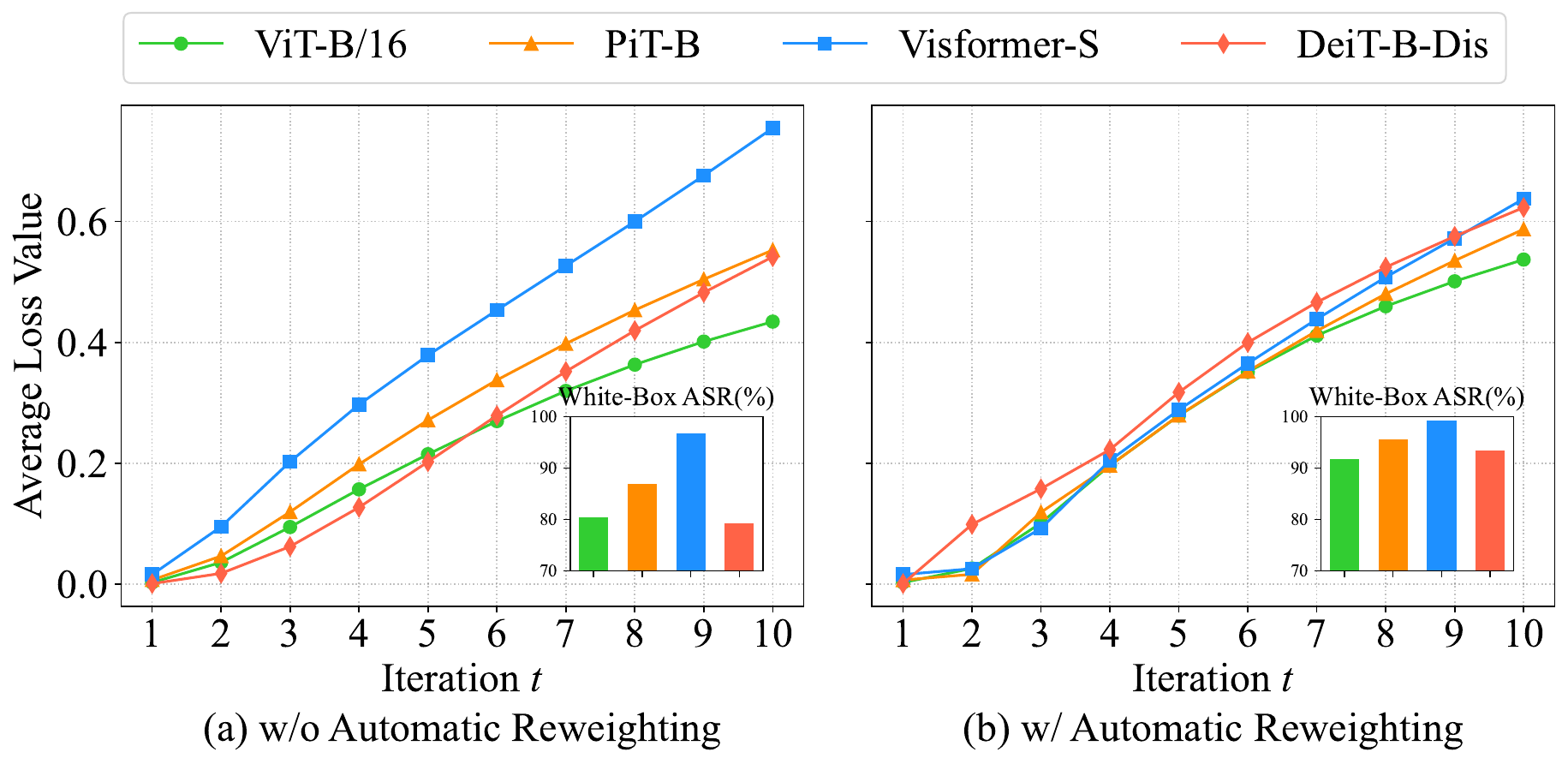}
\caption{Comparison of average loss values during the attack process for ViT-B/16, PiT-B, Visformer-S, and Deit-B-Dis over 10 iterations, (a) without and (b) with Automatic Reweighting, with embedded bar charts showing the final white-box attack success rate (ASR) for each surrogate model.}
\label{w/o reweight}
\end{figure}

\textbf{Automatic Reweighting.}
Due to the difference in inner architecture between surrogate models, the loss calculated on each model will exhibit different magnitudes.
It is more likely that adversarial examples will overfit to the models with larger loss values because they play a more important role in the backpropagation of gradients.
Figure~\ref{w/o reweight} (a) shows when averaging the ensemble weights, Visformer-S has the largest loss value and it also achieves the highest attack success rate of nearly 100\%.
However, models with low loss values, such as ViT-B/16 and DeiT-B-Dis, achieve less than 80\% attack success rate.

To mitigate this issue, we propose an Automatic Reweighting module to balance the contribution of each model to the loss calculation. 
Specifically, we record the loss values of all surrogate models at each iteration and assign weights to each model according to the following equation:
\begin{equation}\label{automatic-reweight} 
w_i=\frac{(\frac{L_{max}}{L_i})^b}{\sum_{j=1}^{N}(\frac{L_{max}}{L_j})^b},
\end{equation}
where $L_{\text{max}} = \max\{L_1, \dots, L_N\}$ is the maximum loss among all surrogate models, $L_i$ denotes the loss of the $i$-th model $f_i$, and $b$ is the hyper-parameter.
Figure \ref{w/o reweight} (b) provides the loss value and attack performance with Automatic Reweighting. 
The results demonstrate that it effectively reduces discrepancy in loss magnitudes across surrogate models and enhances the white-box attack success rate, especially for those with low loss values originally.

\begin{algorithm}[t]
\caption{\name}
\label{\name}
\textbf{Input}: Loss function $J$, surrogate models $\{f_1,...,f_N\}$, a clean image $x$ with ground-truth label $y$, the maximum perturbation $\epsilon$, number of iterations $T$, inference times $loop$, step size enlargement times $q$, momentum decay factor $\mu$ , objective function $OF$, Bayesian optimization function $gp\_minimize$, parameter selection space $P$, the number of calls to the objective function $n_{calls}$.\\
\textbf{Output}: Adversarial images $x^{adv}$.

\begin{algorithmic}[1] 
\STATE \textcolor{darkyellow}{\# Phase 1: Model Augmentation}

\FOR{i=0 to $N$-1}

\STATE Set $F=\{f_1,...,f_{i-1},f_{i+1},...,f_N\}$.
\STATE Build Bayesian optimization process\\ $gp\_minimize(n_{calls},P,OF(p\in P,c,f_i,F))$
\STATE $\tau_i=gp\_minimize (c=MHD)$
\STATE $s_i,\xi_i=gp\_minimize (c=ASS)$
\STATE $\rho_i=gp\_minimize (c=MFM)$
\ENDFOR
\STATE \textcolor{darkyellow}{\# Phase 2: Ensemble Attack}
\STATE Set step size $\alpha = \frac{q\cdot\epsilon}{T} , g_{0} = 0,x_0^{adv}=x $.
\FOR{$t=0$ to $T-1$}

\FOR{$i=0$ to $N-1$}
\FOR{$j=0$ in $loop-1$}
\STATE $L_i=J(f^{MHD}_{\tau_i}(x_{t}^{adv}),y)+J(f^{ASS}_{s_i,\xi_i}(x_{t}^{adv}),y)$
\STATE $\phantom{L_i=}+J(f^{MFM}_{\rho_i}(x_{t}^{adv}),y)$
\ENDFOR
\ENDFOR
\STATE Calculate $\{w_1,...,w_N\}$ using Eq. (\ref{automatic-reweight}).\\
\STATE $g_{t+1}=\nabla_{x_{t}^{adv}} (\sum^{N}_{i=1}w_i\cdot L_i)$
\STATE$g_{t+1}=\mu \cdot g_{t}+\frac{g_{t+1}}{\Vert g_{t+1} \Vert_1 } $\\
\STATE $x_{t+1}^{adv}=x_{t}^{adv}+\alpha \cdot$ sign($g_{t+1}$)\\
\ENDFOR
\RETURN $x^{adv}$\\
\end{algorithmic}
\end{algorithm}

\textbf{Step Size Enlargement.}
Traditionally, the step size $\alpha$ in each iteration is set to $\frac{\epsilon}{T}$, where $\epsilon$ is the maximum perturbation and $T$ is the number of attack iterations.
However, as shown in Figure~\ref{w/o reweight} (a), we find that while using the basic ensemble attack setting (Ens), ensemble models retain a large margin to 100\% white-box attack success rate, indicating that the attack process has not converged yet.
Hence, we propose Step Size Enlargement to enhance the attack strength and accelerate the convergence process.
Specifically, we set the step size as $\alpha=\frac{q\cdot\epsilon}{T}$, and $q$ is the hyper-parameter.
We do comprehensive ablation studies to test the attack performance under different $q$ and validate that a large step size leads to high transferability.\par

\textbf{Overall attack framework.}
We present the details of \name in Algorithm \ref{\name}, and there are two aspects that should be highlighted.
First, to take full advantage of the randomness of our method and improve the diversity of ensemble models, 
we perform inference $loop$ times for the augmented models.
Second, model augmentation and ensemble attack are two independent processes.
Note that the model augmentation is a pre-process that takes only once. 
When generating adversarial examples, most of the time consumption depends on the number of ensemble models and the inference times.
\section{Experiments}
\begin{table*}[ht]
\centering
\begin{tabular}{|c|c|c|c|c|c|c|c|c|c|}
\hline
Attack & Model & CaiT-S/24 & TNT-S & LeViT-256 & ConViT-B & RVT-S$^{*}$ & Drvit & Vit+DAT & ViT-B/16$_{AT}$ \\
\hline
\multirow{5}{*}{I-FGSM} & Ens &63.6 & 60.9 & 48.6 & 61.4 & 47.8 & 59.1 & 50.4 & 97.6 \\ 
        & SVRE & 94.1 & 90.2 & 74.3 & 92.9 & 75.9 & 88.6 & 84.3 & 97.7 \\ 
        & AdaEA & 86.8 & 78.9 & 61.0 & 84.8 & 60.0 & 76.5 & 70.5 & 97.6 \\
        & SMER & 95.3 & 90.4 & 78.6 & 94.1 & 79.7 & 90.0 & 86.4 & 97.8 \\
        & Ours & \textbf{99.1} & \textbf{98.1} & \textbf{95.4} & \textbf{99.0} & \textbf{92.7} & \textbf{97.4} & \textbf{97.1} & \textbf{97.9} \\ 
        \hline
        \multirow{5}{*}{MI-FGSM} & Ens & 76.1 & 74.8 & 69.0 & 74.6 & 69.4 & 72.5 & 69.8 & 97.7 \\
        & SVRE & 99.5 & 97.9 & 95.1 & 99.4 & 94.3 & 97.6 & 97.8 & 97.8 \\
        & AdaEA & 96.4 & 93.7 & 86.3 & 95.9 & 86.8 & 93.8 & 92.4 & 97.8 \\
        & SMER & \textbf{99.7} & 98.1 & 95.0 & \textbf{99.5} & 94.2 & 97.8 & 97.4 & 97.8 \\
        & Ours & 99.5 & \textbf{99.0} & \textbf{98.5} & 99.3 & \textbf{97.3} & \textbf{99.3} & \textbf{99.1} & \textbf{97.9} \\
        \hline
        \multirow{5}{*}{DI-FGSM} & Ens & 78.0 & 78.5 & 73.7 & 76.5 & 72.5 & 74.6 & 69.0 & 97.4 \\
        & SVRE & 98.9 & 98.5 & 96.7 & 98.5 & 95.1 & 97.8 & 96.0 & 97.8 \\
        & AdaEA & 92.1 & 89.9 & 81.0 & 91.2 & 81.5 & 88.6 & 84.9 & 97.5 \\
        & SMER & 99.0 & 98.0 & 96.9 & 98.6 & 96.0 & 98.3 & 96.4 & 97.8 \\
        & Ours & \textbf{99.9} & \textbf{100.0} & \textbf{99.8} & \textbf{100.0} & \textbf{99.7} & \textbf{100.0} & \textbf{99.4} & \textbf{98.0} \\
        \hline
        \multirow{5}{*}{TI-FGSM} & Ens & 70.9 & 68.9 & 55.2 & 68.5 & 55.0 & 67.7 & 58.2 & 97.6 \\
        & SVRE & 94.8 & 92.5 & 79.1 & 93.9 & 81.2 & 92.8 & 87.8 & 97.7 \\
        & AdaEA & 90.2 & 84.9 & 67.1 & 88.5 & 68.1 & 84.4 & 77.5 & 97.8 \\
        & SMER& 95.9 & 93.6 & 81.4 & 94.9 & 83.1 & 93.9 & 89.4 & 97.8 \\ 
        & Ours & \textbf{99.5} & \textbf{99.1} & \textbf{97.8} & \textbf{99.4} & \textbf{95.4} & \textbf{99.2} & \textbf{98.3} & \textbf{97.9} \\
        \hline
\end{tabular}
\caption{The attack success rates (\%) against eight ViTs by various transfer-based ensemble attacks. The best results appear in bold.}
\label{tablevit}
\end{table*}

\begin{table*}[ht]
\centering
\begin{tabular}{|c|c|c|c|c|c|c|c|c|c|}
\hline
Attack & Model & Inc-v3 & Inc-v4 & IncRes-v2 & Res-v2 & Inc-v3\(_{ens3}\) & Inc-v3\(_{ens4}\) & IncRes-v2\(_{adv}\) & MViT-v2\\
\hline
\multirow{5}{*}{I-FGSM} 
& Ens & 38.8 & 38.4 & 32.6 & 34.6 & 26.1 & 23.1 & 17.9 & 30.2 \\
& SVRE & 63.3 & 61.9 & 55.1 & 54.9 & 46.3 & 41.6 & 32.9 & 50.9 \\
& AdaEA & 47.4 & 44.8 & 38.4 & 40.8 & 29.2 & 26.7 & 20.2 & 35.4 \\
& SMER & 64.9 & 62.5 & 57.5 & 58.4 & 48.7 & 46.0 & 37.6 & 53.7 \\
& Ours & \textbf{90.3} & \textbf{88.1} & \textbf{84.5} & \textbf{84.6} & \textbf{76.8} & \textbf{70.7} & \textbf{61.8} & \textbf{79.5} \\
\hline
\multirow{5}{*}{MI-FGSM} 
& Ens & 66.3 & 64.3 & 60.4 & 63.3 & 54.2 & 50.3 & 46.5 & 57.9 \\
& SVRE & 88.4 & 87.2 & 87.4 & 84.6 & 78.0 & 72.5 & 68.5 & 80.9 \\
& AdaEA & 76.5 & 77.3 & 73.4 & 73.0 & 66.9 & 62.4 & 59.0 & 69.8 \\
& SMER & 88.2 & 87.7 & 85.8 & 84.7 & 77.6 & 74.1 & 68.8 & 81.0 \\
& Ours & \textbf{95.7} & \textbf{95.3} & \textbf{94.0} & \textbf{93.3} & \textbf{89.0} & \textbf{84.4} & \textbf{80.0} & \textbf{90.2} \\
\hline
\multirow{5}{*}{DI-FGSM} 
& Ens & 67.3 & 67.1 & 60.9 & 62.1 & 54.1 & 50.9 & 45.9 & 58.3 \\
& SVRE & 91.9 & 92.2 & 90.9 & 87.1 & 82.9 & 80.4 & 76.7 & 86.0 \\
& AdaEA & 70.9 & 70.4 & 64.7 & 63.6 & 57.6 & 53.6 & 47.5 & 61.2 \\
& SMER & 93.4 & 92.7 & 91.1 & 87.7 & 84.1 & 82.0 & 76.8 & 86.8 \\
& Ours & \textbf{99.0} & \textbf{99.2} & \textbf{98.3} & \textbf{97.0} & \textbf{97.2} & \textbf{96.1} & \textbf{93.8} & \textbf{97.2} \\
\hline
\multirow{5}{*}{TI-FGSM} 
& Ens & 46.4 & 45.6 & 39.9 & 40.2 & 31.6 & 29.2 & 23.7 & 36.7 \\
& SVRE & 68.9 & 70.6 & 62.6 & 61.2 & 56.8 & 54.5 & 47.0 & 60.2 \\
& AdaEA & 55.1 & 53.0 & 47.0 & 47.7 & 38.2 & 35.4 & 29.4 & 43.7 \\
& SMER & 73.8 & 71.9 & 64.6 & 63.5 & 59.2 & 56.4 & 50.4 & 62.8 \\
& Ours & \textbf{93.9} & \textbf{94.8} & \textbf{90.5} & \textbf{88.9} & \textbf{84.8} & \textbf{82.2} & \textbf{76.2} & \textbf{87.3} \\
\hline
\end{tabular}
\caption{The attack success rates (\%) against eight CNNs by various transfer-based ensemble attacks. The best results appear in bold.}
\label{tablecnn}
\end{table*}

In this section, we begin by detailing our experimental setup, then compare our method with the latest adversarial ensemble attacks against ViTs and CNNs. 
This comparison highlights the effectiveness of our method in enhancing ensemble transferability between ViTs as well as cross-structure transferability.
We also do ablation studies on the modules of ViT-EnsembleAttack, hyperparameters $q$, $b$, $loop$, and resource consumption.
Finally, we further analyze the effect of each augmentation strategy on the transferability of adversarial examples.\par
\subsection{Experimental Setup}
We compare the performance of ViT-EnsembleAttack with existing state-of-the-art methods against the normally trained ViTs, robust ViTs, adversarially trained ViTs, normally trained CNNs, adversarially trained CNNs, and a hybrid model, respectively. Our experiments concentrate on the image classification task. 

\textbf{Dataset.} We randomly sample 1000 images from the ILSVRC 2012 validation set~\cite{russakovsky2015imagenet} as the clean images to be attacked, then randomly sample another 4000 different images used for Bayesian optimization. 
We check that all of the surrogate and target models achieve almost 100\% classification success rate on the two sampled datasets.\par

\textbf{Models.}
We choose four representative ViT models as the surrogate models to generate adversarial examples, including ViT-B/16~\cite{DBLP:journals/corr/abs-2010-11929}, PiT-B~\cite{heo2021rethinking}, DeiT-B-Dis~\cite{touvron2021training}, and Visformer-S~\cite{chen2021visformer}.
We evaluate the transferability of adversarial examples of ViTs under two attacking scenarios. 
One is that the surrogate and target models are both ViTs to validate the transferability across different ViTs. 
The other is that the surrogate models are ViTs, but the target models are CNNs to examine the cross-model structure transferability. 
For the first setting, the target ViT models contain four normally trained ViTs: CaiT-S/24~\cite{touvron2021going}, TNT-S ~\cite{han2021transformer}, LeViT-256~\cite{graham2021levit}, ConViT-B~\cite{d2021convit}, three robust ViTs: RVT-S$^{*}$~\cite{mao2022towards}, Drvit~\cite{mao2021discrete}, Vit+Dat~\cite{mao2022enhance}, and an adversarially trained ViT: ViT-B/16$_{AT}$~\cite{mo2022adversarial}.
For the second setting, we select normally trained CNNs: Inception-v3 (Inc-v3)~\cite{szegedy2016rethinking}, Inception-v4 (Inc-v4)~\cite{szegedy2017inception}, Inception-Resnet-v2 (IncRes-v2)~\cite{szegedy2017inception}, Resnet-v2-152 (Res-v2)~\cite{he2016deep}, adversarially trained models: an ensemble of three adversarial trained Inceptionv3 models (Inc-v3$_{ens3}$)~\cite{tramer2017ensemble}, an ensemble of four adversarial trained Inception-v3 models (Inc-v3$_{ens4}$)~\cite{tramer2017ensemble}, adversarial trained Inception-Resnet-v2 (IncRes-v2$_{adv}$)~\cite{tramer2017ensemble} and a hybrid model MobileViTv2 (MViTv2)~\cite{mehta2022separable} which has both convolutional layers and ViT blocks as the target models. \par 

\textbf{Comparisons and baselines.} 
We choose the ensemble attack (Ens), which updates adversarial examples using Eq~(\ref{Ensemble-I-FGSM}) and average weights, and three SOTA methods, SVRE~\cite{xiong2022stochastic}, AdaEA~\cite{chen2023adaptive} and SMER~\cite{tang2024ensemble}, as the competitive baselines. 
All methods are integrated into four attack settings, including I-FGSM~\cite{kurakin2018adversarial}, MI-FGSM~\cite{dong2018boosting}, DI-FGSM~\cite{xie2019improving}, and TI-FGSM~\cite{dong2019evading}.
\par
\textbf{Evaluation metric.} The evaluation metric is the attack success rate (ASR), the ratio of the adversarial examples that successfully mislead the target model among all samples.\par
\textbf{Hyper-parameters.} For a fair comparison, we follow the hyper-parameters setting in ~\cite{tang2024ensemble} to set the maximum perturbation to $\epsilon = 16$ and the number of iterations to $T = 10$, so the step size in other methods is $\alpha = \frac{\epsilon}{T} = 1.6$. 
Hyper-parameters of other methods follow their default settings.
For the decay factor $\mu$ in MI-FGSM, we set $\mu$ to 1.0. 
For the translation kernel in TI-FGSM, we use the Gaussian kernel, the size is $5\times5$.
For transformation operation $T(\cdot;p)$ in DI-FGSM, we set $p=0.5$ and the range of $rnd$ is $[224,248)$. We set $n_{calls}=50$, $P=(0,1)$ for $gp\_minimize$ function.  
For the other hyper-parameters in ViT-EnsembleAttack, we set $loop=2$, $q=3$ and $b=2$.
All images are resized to 224 × 224 to conduct experiments and set the patch size to 16 for the inputs of ViTs. 
\par
\subsection{Transferability}
Here we analyze the performance of our approach against ViTs and CNNs, respectively. Specifically, we generate adversarial examples on four given surrogate models and directly attack various target models to show the generalization of the proposed method. \par
\textbf{Performance on ViTs.} 
We first compare the general attack performance of \name with existing ensemble methods on the normally trained, robust and adversarially trained ViTs.
As shown in Table \ref{tablevit}, in the black-box setting, our method outperforms the state-of-the-art baselines by a large average margin of 4.6\% attack success rate on average.
Specifically, our method improves the attack success rate from 78.6\% to 95.4\% on LeViT-256 when integrating with I-FGSM. 
For DI-FGSM, our method achieves an attack success rate of nearly 100\%, further demonstrating its effectiveness.\par
\textbf{Performance on CNNs.}  
We then attempt to evaluate the cross-structure transferability by attacking normally trained and adversarially trained CNNs.
The results are summarized in Table~\ref{tablecnn}.
It can be seen that the attack success rate decreases a lot compared to attacking ViTs, illustrating the difficulty of cross-model structure transfer attack.
Nevertheless, our method still achieves nearly 88.3\% attack success rate on average, outperforming SMER by a significant margin of 15.3\% on average, which represents a substantial advancement over prior methods, demonstrating the superior cross-structure transferability performance of our proposed \name.\par
\subsection{Ablation Study}
\label{ablationstudy}
In this subsection, we analyze the contribution of each module and study the effects of several key hyper-parameters to justify our choices.\par
\begin{table}[t!]
    \centering
    \setlength{\tabcolsep}{1.1mm}{
    \begin{tabular}{|ccc|cc|}
        \hline
        Augmentation & Reweighting & Enlargement & ViTs & CNNs\\
        \hline
        - & - & - & 70.5 & 48.1 \\
        \checkmark & - & - & 93.4 & 78.8 \\
        - & \checkmark & - & 78.6 & 52.6\\
        - & - & \checkmark & 87.1 & 65.5 \\
        \checkmark & \checkmark & - & 95.7 & 81.0  \\
        \checkmark & - & \checkmark & 98.0 & 88.1  \\
        - & \checkmark & \checkmark & 89.2 & 66.9  \\
        \checkmark & \checkmark & \checkmark & 98.4 & 88.3 \\
        \hline
    \end{tabular}}
    \caption{The average attack success rates (\%) against ViTs and CNNs by various settings of modules. \checkmark indicates that the module is applied. For simplicity, we only retain the last word of each module.}
    \label{modules}
\end{table}
\textbf{On the modules of ViT-EnsembleAttack.} We integrate our method with all attack algorithms, utilizing various modules to craft adversarial examples, and report their transferability on ViTs and CNNs.
As shown in Table~\ref{modules}, Model Augmentation module improves the attack success rate mostly, indicating its effectiveness in ViT-based ensemble attacks.
Automatic Reweighting and Step Size Enlargement each surpass the baseline individually, and their combination outperforms either alone.
When paired with augmentation, both techniques improve upon augmentation alone, with the best results achieved by combining all three, exceeding any single or pairwise setup.
This outcome demonstrates that the three modules in \name are complement and combine each other could achieve the improvement of transferability. \par
\begin{figure}[t]
\centering
\includegraphics[width=1\columnwidth]{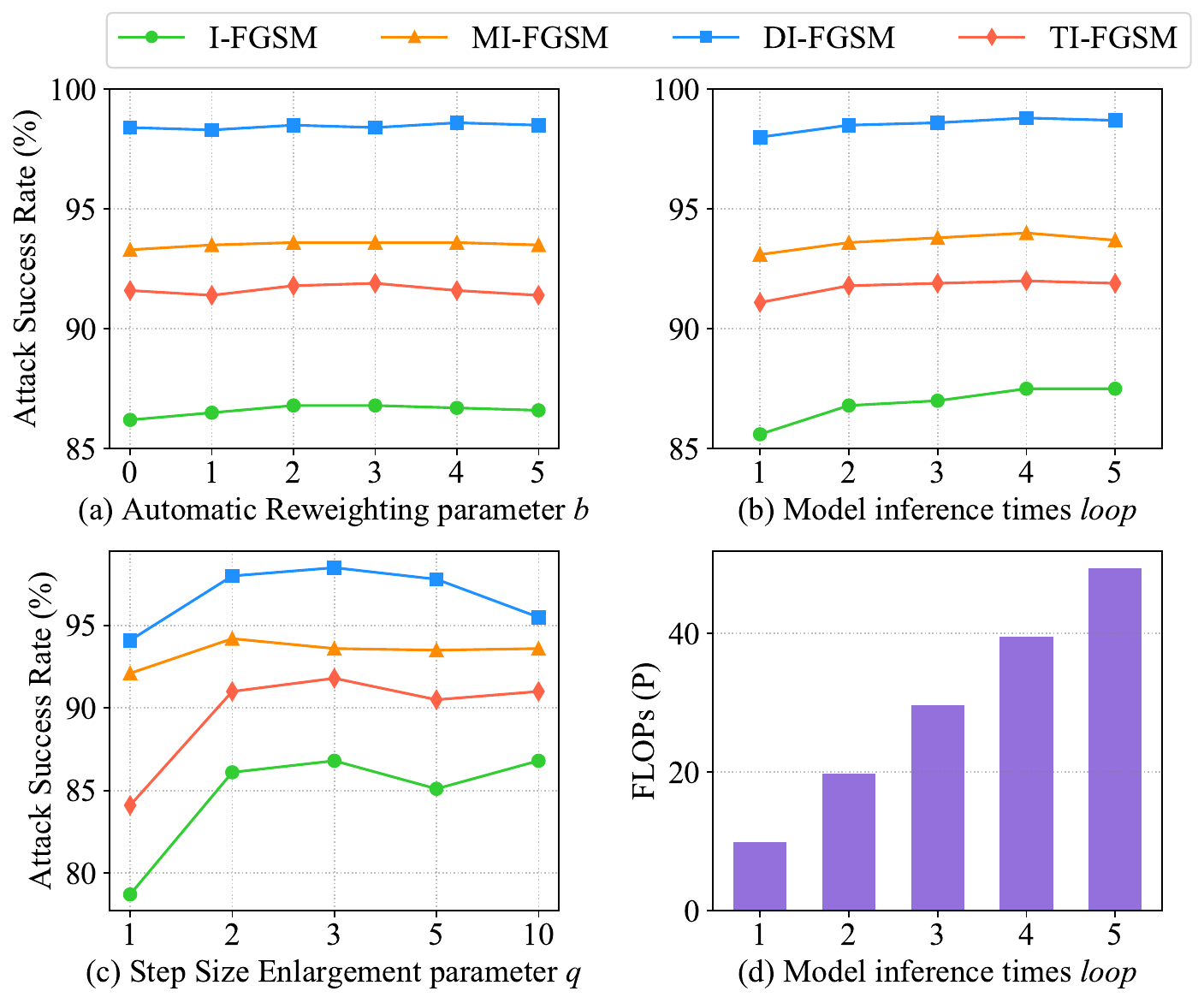}
\caption{Average attack success rate against ViTs and CNNs under three varying parameters: (a) automatic reweighting parameter $b$, (b) model inference times $loop$, and (c) step size enlargement parameter $q$. (d) Computational cost (FLOPs) for different model inference times $loop$.}
\label{bqloop}
\end{figure}
\textbf{On hyper-parameter sensitivity.}
We conduct a detailed analysis of the key hyper-parameters $b$, $q$, and $loop$ to explain the optimal configuration.
As shown in Figure~\ref{bqloop} (a),  the variation in attack success rate with changes in $b$, except for $b=0$, is not significant.
We set $b=2$ as the final choice because it maintains high attack success rates across all algorithms, making it a balanced option.
Figure~\ref{bqloop} (c) illustrates that a moderate increase in $q$ enhances attack success, with the peak performance observed at $q=3$ for most algorithms. 
However, beyond this point (\eg, $q=5$ and $q=10$), the attack success rate declines, likely due to instability caused by excessively large step sizes. 
Based on this observation, we select $q=3$ as the optimal value.
Figure~\ref{bqloop} (b) exhibits that increasing $loop$ improves the attack success rate, but the gains become marginal beyond $loop=2$. Meanwhile, Figure~\ref{bqloop} (d) indicates that the computational cost grows exponentially with larger $loop$ values.
Given the trade-off between attack effectiveness and computational efficiency, we choose $loop=2$ to balance performance and resource consumption.
\par
Increasing the number of $loop$ iterations improves attack success because our method uses model augmentation to inject randomness during inference, resulting in varied gradient estimates at each back-propagation.
Accumulating these diverse directions over multiple rounds enhances transferability. 
Without model augmentation, repeated inference yields identical gradients. Thus, $loop$ is designed to amplify the effect of model augmentation.
\par
\begin{table}[h]
\centering
\footnotesize 
\caption{Computational resource consumption of different methods. We report the result of our method into two phases, as  described in Algorithm~\ref{\name}.}
\label{tab:resource_comparison}
    \begin{tabularx}{\columnwidth}{|p{1.3cm}|XXXX|XX|} %
        \hline
        & \multirow{2}{*}{Ens} & \multirow{2}{*}{SVRE} & \multirow{2}{*}{AdaEA} & \multirow{2}{*}{SMER} & \multicolumn{2}{c|}{Ours}\\
        \cline{6-7}&&&&& Phase1 & Phase2 \\
        \hline
        FLOPs (P) & 3.290 & 29.623 & 18.653 & 61.309 & 54.069 & 19.738 \\
        Time (s)  & 395.2  & 3460.2  & 2176.0  & 7573.9  & 2394.7  & 2669.9  \\
        \hline
    \end{tabularx}
\end{table}
\textbf{On resource consumption.}
In Table~\ref{tab:resource_comparison}, we report both floating-point operations per second (FLOPs) and time to compare computational resource consumption of all methods.
Since our method includes two phases, we calculate the resource consumption on the two phases separately.
Our method consumes 54.069P FLOPs and takes 2394.7 seconds in Phase 1. 
Although the resource consumption in Phase 1 is relatively high, it is worth noting that Phase 1 only needs to be executed once. 
In Phase 2, our method consumes 19.738P FLOPs and takes 2669.9 seconds. Compared to Phase 1, the resource consumption in Phase 2 is significantly reduced. 
Compared to other methods, such as SMER and SVRE, our method consumes fewer resources in general during the attack process. \par
\begin{figure}[t]
\centering
\includegraphics[width=1\columnwidth]{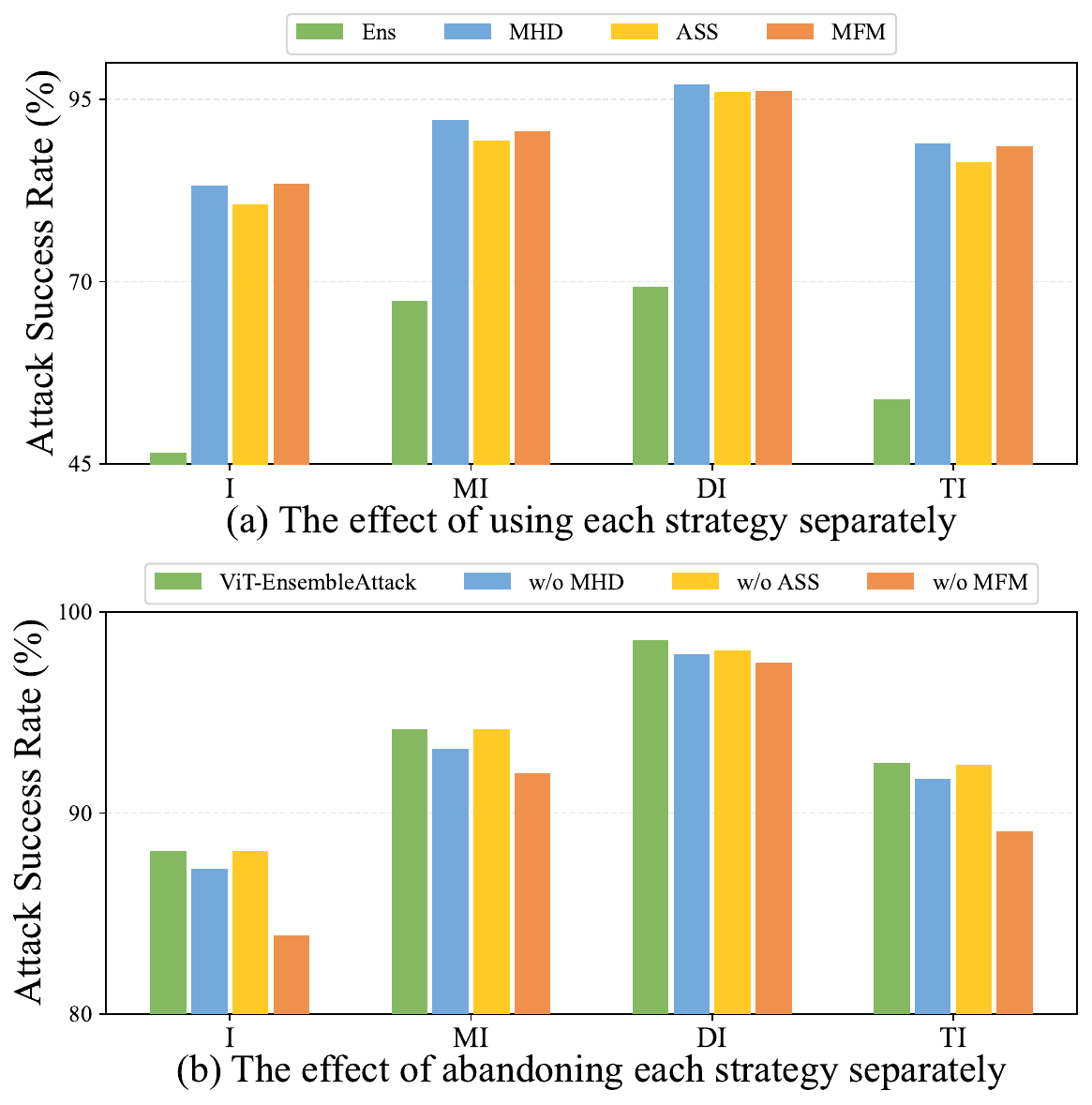}
\caption{The average attack success rates (\%) against ViTs and CNNs with different settings of augment strategies: (a) the effect of using each strategy separately, (b) the effect of abandoning each strategy separately.}
\label{augment}
\end{figure}
\subsection{Further Analysis}
Since we design three strategies for model augmentation, we further analyze the effect of each strategy on the transferability of adversarial examples.\par
\textbf{Whether each strategy contributes to the improvement of transferability?} 
We first conduct experiments to test the attack performance when using the three strategies separately.
From Figure~\ref{augment}(a), it can be observed that all three strategies significantly improve the attack success rate over the Ens setting, demonstrating their effectiveness in augmenting the surrogate models. \par
\textbf{Is each strategy indispensable to the overall attack performance?}
We further conduct experiments to test the effect of abandoning each strategy on the overall attack success rate.
It can be seen from Figure~\ref{augment} (b) that when abandoning one strategy, the attack success rate declines in most cases, demonstrating that each strategy is indispensable in our model augmentation.
We also observe an interesting phenomenon: when abandoning MFM, the attack success rate declines the most.
We believe this is because MHD and ASS are both designed for the multi-head attention module, restricting the diversity of augmented models.
In contrast, when abandoning MHD or ASS, the remaining two strategies are for multi-head attention and multi-layer perception, ensuring diversity and achieving higher performance.
\section{Conclusion}
In this work, we propose ViT-EnsembleAttack, a novel ensemble-based adversarial attack designed for ViTs. 
Different from prior ensemble-based attacks, we propose to augment surrogate models by increasing diversity to enhance the transferability of adversarial examples.
Extensive experimental results 
show that our method outperforms state-of-the-art methods by a substantial margin across various transfer settings. 
The core innovation of our method lies in the adversarial augmentation of the surrogate models.
Future work could explore new augmentation techniques on ViTs and other kinds of models to enhance the ensemble-based adversarial transferability.
\section*{Acknowledgments}
This work is supported by the National Natural Science Foundation (U22B2017) and the International Cooperation Foundation of Hubei Province, China (2024EHA032).
{
    \small
    \bibliographystyle{ieeenat_fullname}
    \bibliography{main}

\begin{thebibliography}{48}
\providecommand{\natexlab}[1]{#1}
\providecommand{\url}[1]{\texttt{#1}}
\expandafter\ifx\csname urlstyle\endcsname\relax
  \providecommand{\doi}[1]{doi: #1}\else
  \providecommand{\doi}{doi: \begingroup \urlstyle{rm}\Url}\fi

\bibitem[Chen et~al.(2023)Chen, Yin, Chen, Chen, and Liu]{chen2023adaptive}
Bin Chen, Jiali Yin, Shukai Chen, Bohao Chen, and Ximeng Liu.
\newblock An adaptive model ensemble adversarial attack for boosting adversarial transferability.
\newblock In \emph{Proceedings of the IEEE/CVF International Conference on Computer Vision}, pages 4489--4498, 2023.

\bibitem[Chen et~al.(2021)Chen, Xie, Niu, Liu, Wei, and Tian]{chen2021visformer}
Zhengsu Chen, Lingxi Xie, Jianwei Niu, Xuefeng Liu, Longhui Wei, and Qi Tian.
\newblock Visformer: The vision-friendly transformer.
\newblock In \emph{Proceedings of the IEEE/CVF international conference on computer vision}, pages 589--598, 2021.

\bibitem[Dong et~al.(2018)Dong, Liao, Pang, Su, Zhu, Hu, and Li]{dong2018boosting}
Yinpeng Dong, Fangzhou Liao, Tianyu Pang, Hang Su, Jun Zhu, Xiaolin Hu, and Jianguo Li.
\newblock Boosting adversarial attacks with momentum.
\newblock In \emph{Proceedings of the IEEE conference on computer vision and pattern recognition}, pages 9185--9193, 2018.

\bibitem[Dong et~al.(2019)Dong, Pang, Su, and Zhu]{dong2019evading}
Yinpeng Dong, Tianyu Pang, Hang Su, and Jun Zhu.
\newblock Evading defenses to transferable adversarial examples by translation-invariant attacks.
\newblock In \emph{Proceedings of the IEEE/CVF conference on computer vision and pattern recognition}, pages 4312--4321, 2019.

\bibitem[Dosovitskiy et~al.(2020)Dosovitskiy, Beyer, Kolesnikov, Weissenborn, Zhai, Unterthiner, Dehghani, Minderer, Heigold, Gelly, Uszkoreit, and Houlsby]{DBLP:journals/corr/abs-2010-11929}
Alexey Dosovitskiy, Lucas Beyer, Alexander Kolesnikov, Dirk Weissenborn, Xiaohua Zhai, Thomas Unterthiner, Mostafa Dehghani, Matthias Minderer, Georg Heigold, Sylvain Gelly, Jakob Uszkoreit, and Neil Houlsby.
\newblock An image is worth 16x16 words: Transformers for image recognition at scale.
\newblock \emph{CoRR}, abs/2010.11929, 2020.

\bibitem[d’Ascoli et~al.(2021)d’Ascoli, Touvron, Leavitt, Morcos, Biroli, and Sagun]{d2021convit}
St{\'e}phane d’Ascoli, Hugo Touvron, Matthew~L Leavitt, Ari~S Morcos, Giulio Biroli, and Levent Sagun.
\newblock Convit: Improving vision transformers with soft convolutional inductive biases.
\newblock In \emph{International conference on machine learning}, pages 2286--2296. PMLR, 2021.

\bibitem[Ganeshan et~al.(2019)Ganeshan, BS, and Babu]{ganeshan2019fda}
Aditya Ganeshan, Vivek BS, and R~Venkatesh Babu.
\newblock Fda: Feature disruptive attack.
\newblock In \emph{Proceedings of the IEEE/CVF International Conference on Computer Vision}, pages 8069--8079, 2019.

\bibitem[Ge et~al.(2023{\natexlab{a}})Ge, Liu, Wang, Shang, and Liu]{ge2023boosting}
Zhijin Ge, Hongying Liu, Xiaosen Wang, Fanhua Shang, and Yuanyuan Liu.
\newblock {Boosting Adversarial Transferability by Achieving Flat Local Maxima}.
\newblock In \emph{Proceedings of the Advances in Neural Information Processing Systems}, 2023{\natexlab{a}}.

\bibitem[Ge et~al.(2023{\natexlab{b}})Ge, Shang, Liu, Liu, Wan, Feng, and Wang]{ge2023improving}
Zhijin Ge, Fanhua Shang, Hongying Liu, Yuanyuan Liu, Liang Wan, Wei Feng, and Xiaosen Wang.
\newblock {Improving the Transferability of Adversarial Examples with Arbitrary Style Transfer}.
\newblock In \emph{Proceedings of the ACM International Conference on Multimedia}, page 4440–4449, 2023{\natexlab{b}}.

\bibitem[Goodfellow et~al.(2014)Goodfellow, Shlens, and Szegedy]{goodfellow2014explaining}
Ian~J Goodfellow, Jonathon Shlens, and Christian Szegedy.
\newblock Explaining and harnessing adversarial examples.
\newblock \emph{arXiv preprint arXiv:1412.6572}, 2014.

\bibitem[Graham et~al.(2021)Graham, El-Nouby, Touvron, Stock, Joulin, J{\'e}gou, and Douze]{graham2021levit}
Benjamin Graham, Alaaeldin El-Nouby, Hugo Touvron, Pierre Stock, Armand Joulin, Herv{\'e} J{\'e}gou, and Matthijs Douze.
\newblock Levit: a vision transformer in convnet's clothing for faster inference.
\newblock In \emph{Proceedings of the IEEE/CVF international conference on computer vision}, pages 12259--12269, 2021.

\bibitem[Guo et~al.(2017)Guo, Rana, Cisse, and Van Der~Maaten]{guo2017countering}
Chuan Guo, Mayank Rana, Moustapha Cisse, and Laurens Van Der~Maaten.
\newblock Countering adversarial images using input transformations.
\newblock \emph{arXiv preprint arXiv:1711.00117}, 2017.

\bibitem[Han et~al.(2021)Han, Xiao, Wu, Guo, Xu, and Wang]{han2021transformer}
Kai Han, An Xiao, Enhua Wu, Jianyuan Guo, Chunjing Xu, and Yunhe Wang.
\newblock Transformer in transformer.
\newblock \emph{Advances in neural information processing systems}, 34:\penalty0 15908--15919, 2021.

\bibitem[He et~al.(2016)He, Zhang, Ren, and Sun]{he2016deep}
Kaiming He, Xiangyu Zhang, Shaoqing Ren, and Jian Sun.
\newblock Deep residual learning for image recognition.
\newblock In \emph{Proceedings of the IEEE conference on computer vision and pattern recognition}, pages 770--778, 2016.

\bibitem[Heo et~al.(2021)Heo, Yun, Han, Chun, Choe, and Oh]{heo2021rethinking}
Byeongho Heo, Sangdoo Yun, Dongyoon Han, Sanghyuk Chun, Junsuk Choe, and Seong~Joon Oh.
\newblock Rethinking spatial dimensions of vision transformers.
\newblock In \emph{Proceedings of the IEEE/CVF international conference on computer vision}, pages 11936--11945, 2021.

\bibitem[Johnson and Zhang(2013)]{johnson2013accelerating}
Rie Johnson and Tong Zhang.
\newblock Accelerating stochastic gradient descent using predictive variance reduction.
\newblock \emph{Advances in neural information processing systems}, 26, 2013.

\bibitem[Kurakin et~al.(2018)Kurakin, Goodfellow, and Bengio]{kurakin2018adversarial}
Alexey Kurakin, Ian~J Goodfellow, and Samy Bengio.
\newblock Adversarial examples in the physical world.
\newblock In \emph{Artificial intelligence safety and security}, pages 99--112. Chapman and Hall/CRC, 2018.

\bibitem[Li et~al.(2024)Li, Guo, Zuo, and Chen]{li2024improving}
Qizhang Li, Yiwen Guo, Wangmeng Zuo, and Hao Chen.
\newblock Improving adversarial transferability via intermediate-level perturbation decay.
\newblock \emph{Advances in Neural Information Processing Systems}, 36, 2024.

\bibitem[Li et~al.(2020)Li, Bai, Zhou, Xie, Zhang, and Yuille]{li2020learning}
Yingwei Li, Song Bai, Yuyin Zhou, Cihang Xie, Zhishuai Zhang, and Alan Yuille.
\newblock Learning transferable adversarial examples via ghost networks.
\newblock In \emph{Proceedings of the AAAI conference on artificial intelligence}, pages 11458--11465, 2020.

\bibitem[Lin et~al.(2019)Lin, Song, He, Wang, and Hopcroft]{lin2019nesterov}
Jiadong Lin, Chuanbiao Song, Kun He, Liwei Wang, and John~E Hopcroft.
\newblock Nesterov accelerated gradient and scale invariance for adversarial attacks.
\newblock \emph{arXiv preprint arXiv:1908.06281}, 2019.

\bibitem[Lin et~al.(2024)Lin, Luo, Niu, He, Xie, Hou, Shen, and Song]{lin2024boosting}
Qinliang Lin, Cheng Luo, Zenghao Niu, Xilin He, Weicheng Xie, Yuanbo Hou, Linlin Shen, and Siyang Song.
\newblock Boosting adversarial transferability across model genus by deformation-constrained warping.
\newblock In \emph{Proceedings of the AAAI Conference on Artificial Intelligence}, pages 3459--3467, 2024.

\bibitem[Liu et~al.(2016)Liu, Chen, Liu, and Song]{liu2016delving}
Yanpei Liu, Xinyun Chen, Chang Liu, and Dawn Song.
\newblock Delving into transferable adversarial examples and black-box attacks.
\newblock \emph{arXiv preprint arXiv:1611.02770}, 2016.

\bibitem[Mao et~al.(2021)Mao, Jiang, Dehghani, Vondrick, Sukthankar, and Essa]{mao2021discrete}
Chengzhi Mao, Lu Jiang, Mostafa Dehghani, Carl Vondrick, Rahul Sukthankar, and Irfan Essa.
\newblock Discrete representations strengthen vision transformer robustness.
\newblock \emph{arXiv preprint arXiv:2111.10493}, 2021.

\bibitem[Mao et~al.(2022{\natexlab{a}})Mao, Chen, Duan, Zhu, Qi, Li, Zhang, Xue, et~al.]{mao2022enhance}
Xiaofeng Mao, Yuefeng Chen, Ranjie Duan, Yao Zhu, Gege Qi, Xiaodan Li, Rong Zhang, Hui Xue, et~al.
\newblock Enhance the visual representation via discrete adversarial training.
\newblock \emph{Advances in Neural Information Processing Systems}, 35:\penalty0 7520--7533, 2022{\natexlab{a}}.

\bibitem[Mao et~al.(2022{\natexlab{b}})Mao, Qi, Chen, Li, Duan, Ye, He, and Xue]{mao2022towards}
Xiaofeng Mao, Gege Qi, Yuefeng Chen, Xiaodan Li, Ranjie Duan, Shaokai Ye, Yuan He, and Hui Xue.
\newblock Towards robust vision transformer.
\newblock In \emph{Proceedings of the IEEE/CVF conference on Computer Vision and Pattern Recognition}, pages 12042--12051, 2022{\natexlab{b}}.

\bibitem[Mehta and Rastegari(2022)]{mehta2022separable}
Sachin Mehta and Mohammad Rastegari.
\newblock Separable self-attention for mobile vision transformers.
\newblock \emph{arXiv preprint arXiv:2206.02680}, 2022.

\bibitem[Mo et~al.(2022)Mo, Wu, Wang, Guo, and Wang]{mo2022adversarial}
Yichuan Mo, Dongxian Wu, Yifei Wang, Yiwen Guo, and Yisen Wang.
\newblock When adversarial training meets vision transformers: Recipes from training to architecture.
\newblock \emph{Advances in Neural Information Processing Systems}, 35:\penalty0 18599--18611, 2022.

\bibitem[Naseer et~al.(2020)Naseer, Khan, Hayat, Khan, and Porikli]{naseer2020self}
Muzammal Naseer, Salman Khan, Munawar Hayat, Fahad~Shahbaz Khan, and Fatih Porikli.
\newblock A self-supervised approach for adversarial robustness.
\newblock In \emph{Proceedings of the IEEE/CVF Conference on Computer Vision and Pattern Recognition}, pages 262--271, 2020.

\bibitem[Russakovsky et~al.(2015)Russakovsky, Deng, Su, Krause, Satheesh, Ma, Huang, Karpathy, Khosla, Bernstein, et~al.]{russakovsky2015imagenet}
Olga Russakovsky, Jia Deng, Hao Su, Jonathan Krause, Sanjeev Satheesh, Sean Ma, Zhiheng Huang, Andrej Karpathy, Aditya Khosla, Michael Bernstein, et~al.
\newblock Imagenet large scale visual recognition challenge.
\newblock \emph{International journal of computer vision}, 115:\penalty0 211--252, 2015.

\bibitem[Szegedy et~al.(2016)Szegedy, Vanhoucke, Ioffe, Shlens, and Wojna]{szegedy2016rethinking}
Christian Szegedy, Vincent Vanhoucke, Sergey Ioffe, Jon Shlens, and Zbigniew Wojna.
\newblock Rethinking the inception architecture for computer vision.
\newblock In \emph{Proceedings of the IEEE conference on computer vision and pattern recognition}, pages 2818--2826, 2016.

\bibitem[Szegedy et~al.(2017)Szegedy, Ioffe, Vanhoucke, and Alemi]{szegedy2017inception}
Christian Szegedy, Sergey Ioffe, Vincent Vanhoucke, and Alexander Alemi.
\newblock Inception-v4, inception-resnet and the impact of residual connections on learning.
\newblock In \emph{Proceedings of the AAAI conference on artificial intelligence}, 2017.

\bibitem[Tang et~al.(2024)Tang, Wang, Bin, Dou, Yang, and Shen]{tang2024ensemble}
Bowen Tang, Zheng Wang, Yi Bin, Qi Dou, Yang Yang, and Heng~Tao Shen.
\newblock Ensemble diversity facilitates adversarial transferability.
\newblock In \emph{Proceedings of the IEEE/CVF Conference on Computer Vision and Pattern Recognition}, pages 24377--24386, 2024.

\bibitem[Touvron et~al.(2021{\natexlab{a}})Touvron, Cord, Douze, Massa, Sablayrolles, and J{\'e}gou]{touvron2021training}
Hugo Touvron, Matthieu Cord, Matthijs Douze, Francisco Massa, Alexandre Sablayrolles, and Herv{\'e} J{\'e}gou.
\newblock Training data-efficient image transformers \& distillation through attention.
\newblock In \emph{International conference on machine learning}, pages 10347--10357. PMLR, 2021{\natexlab{a}}.

\bibitem[Touvron et~al.(2021{\natexlab{b}})Touvron, Cord, Sablayrolles, Synnaeve, and J{\'e}gou]{touvron2021going}
Hugo Touvron, Matthieu Cord, Alexandre Sablayrolles, Gabriel Synnaeve, and Herv{\'e} J{\'e}gou.
\newblock Going deeper with image transformers.
\newblock In \emph{Proceedings of the IEEE/CVF international conference on computer vision}, pages 32--42, 2021{\natexlab{b}}.

\bibitem[Tram{\`e}r et~al.(2017)Tram{\`e}r, Kurakin, Papernot, Goodfellow, Boneh, and McDaniel]{tramer2017ensemble}
Florian Tram{\`e}r, Alexey Kurakin, Nicolas Papernot, Ian Goodfellow, Dan Boneh, and Patrick McDaniel.
\newblock Ensemble adversarial training: Attacks and defenses.
\newblock \emph{arXiv preprint arXiv:1705.07204}, 2017.

\bibitem[Wang et~al.(2024)Wang, He, Wang, and Wang]{wang2024boosting}
Kunyu Wang, Xuanran He, Wenxuan Wang, and Xiaosen Wang.
\newblock Boosting adversarial transferability by block shuffle and rotation.
\newblock In \emph{Proceedings of the IEEE/CVF Conference on Computer Vision and Pattern Recognition}, pages 24336--24346, 2024.

\bibitem[Wang and He(2021)]{wang2021enhancing}
Xiaosen Wang and Kun He.
\newblock {Enhancing the Transferability of Adversarial Attacks through Variance Tuning}.
\newblock In \emph{Proceedings of the IEEE/CVF Conference on Computer Vision and Pattern Recognition}, pages 1924--1933, 2021.

\bibitem[Wang et~al.(2021)Wang, Lin, Hu, Wang, and He]{wang2021boosting}
Xiaosen Wang, Jiadong Lin, Han Hu, Jingdong Wang, and Kun He.
\newblock {Boosting Adversarial Transferability through Enhanced Momentum}.
\newblock In \emph{Proceedings of the British Machine Vision Conference}, 2021.

\bibitem[Wang et~al.(2023{\natexlab{a}})Wang, Zhang, and Zhang]{wang2023structure}
Xiaosen Wang, Zeliang Zhang, and Jianping Zhang.
\newblock {Structure Invariant Transformation for better Adversarial Transferability}.
\newblock In \emph{Proceedings of the IEEE/CVF International Conference on Computer Vision}, pages 4607--4619, 2023{\natexlab{a}}.

\bibitem[Wang et~al.(2023{\natexlab{b}})Wang, Pang, Du, Lin, Liu, and Yan]{wang2023better}
Zekai Wang, Tianyu Pang, Chao Du, Min Lin, Weiwei Liu, and Shuicheng Yan.
\newblock Better diffusion models further improve adversarial training.
\newblock In \emph{International Conference on Machine Learning}, pages 36246--36263. PMLR, 2023{\natexlab{b}}.

\bibitem[Wei et~al.(2022)Wei, Chen, Goldblum, Wu, Goldstein, and Jiang]{wei2022towards}
Zhipeng Wei, Jingjing Chen, Micah Goldblum, Zuxuan Wu, Tom Goldstein, and Yu-Gang Jiang.
\newblock Towards transferable adversarial attacks on vision transformers.
\newblock In \emph{Proceedings of the AAAI Conference on Artificial Intelligence}, pages 2668--2676, 2022.

\bibitem[Xiaosen et~al.(2023)Xiaosen, Tong, and He]{xiaosen2023rethinking}
Wang Xiaosen, Kangheng Tong, and Kun He.
\newblock Rethinking the backward propagation for adversarial transferability.
\newblock \emph{Advances in Neural Information Processing Systems}, 36:\penalty0 1905--1922, 2023.

\bibitem[Xie et~al.(2017)Xie, Wang, Zhang, Ren, and Yuille]{xie2017mitigating}
Cihang Xie, Jianyu Wang, Zhishuai Zhang, Zhou Ren, and Alan Yuille.
\newblock Mitigating adversarial effects through randomization.
\newblock \emph{arXiv preprint arXiv:1711.01991}, 2017.

\bibitem[Xie et~al.(2019)Xie, Zhang, Zhou, Bai, Wang, Ren, and Yuille]{xie2019improving}
Cihang Xie, Zhishuai Zhang, Yuyin Zhou, Song Bai, Jianyu Wang, Zhou Ren, and Alan~L Yuille.
\newblock Improving transferability of adversarial examples with input diversity.
\newblock In \emph{Proceedings of the IEEE/CVF conference on computer vision and pattern recognition}, pages 2730--2739, 2019.

\bibitem[Xiong et~al.(2022)Xiong, Lin, Zhang, Hopcroft, and He]{xiong2022stochastic}
Yifeng Xiong, Jiadong Lin, Min Zhang, John~E Hopcroft, and Kun He.
\newblock Stochastic variance reduced ensemble adversarial attack for boosting the adversarial transferability.
\newblock In \emph{Proceedings of the IEEE/CVF conference on computer vision and pattern recognition}, pages 14983--14992, 2022.

\bibitem[Zhang et~al.(2022)Zhang, Wu, Huang, Huang, Wang, Su, and Lyu]{zhang2022improving}
Jianping Zhang, Weibin Wu, Jen-tse Huang, Yizhan Huang, Wenxuan Wang, Yuxin Su, and Michael~R Lyu.
\newblock Improving adversarial transferability via neuron attribution-based attacks.
\newblock In \emph{Proceedings of the IEEE/CVF Conference on Computer Vision and Pattern Recognition}, pages 14993--15002, 2022.

\bibitem[Zhang et~al.(2023)Zhang, tse Huang, Wang, Li, Wu, Wang, Su, and Lyu]{zhang2023improving}
Jianping Zhang, Jen tse Huang, Wenxuan Wang, Yichen Li, Weibin Wu, Xiaosen Wang, Yuxin Su, and Michael~R. Lyu.
\newblock {Improving the Transferability of Adversarial Samples by Path-Augmented Method}.
\newblock In \emph{Proceedings of the IEEE/CVF Conference on Computer Vision and Pattern Recognition}, pages 8173--8182, 2023.

\bibitem[Zhang et~al.(2024)Zhang, Zhu, Yao, Wang, and Xu]{zhang2024bag}
Zeliang Zhang, Rongyi Zhu, Wei Yao, Xiaosen Wang, and Chenliang Xu.
\newblock {Bag of Tricks to Boost Adversarial Transferability}.
\newblock 2024.

\end{thebibliography}
}

\end{document}